\documentclass[10pt,twocolumn,letterpaper]{article}

\usepackage{ijcb}
\usepackage{times}
\usepackage{epsfig}
\usepackage{graphicx}
\usepackage{amsmath}
\usepackage{amssymb}
\usepackage{url}
\usepackage[caption=false, labelformat=simple]{subfig}

\usepackage{float}
\usepackage[norule,symbol,perpage]{footmisc}

\newcommand*{\affmark}[1][*]{\textsuperscript{#1}}

\ijcbfinalcopy 

\ifijcbfinal\fi

\makeatletter
\def\ps@IEEEtitlepagestyle{
\def\@oddfoot{\mycopyrightnotice}
\def\@evenfoot{}
}
\def\mycopyrightnotice{
{\hfill \footnotesize 978-1-7281-9186-7/20/\$31.00 \copyright 2020 IEEE\hfill}
}
\makeatother

\begin{document}

\title{Pixel Sampling for Style Preserving Face Pose Editing}
\author{
Xiangnan Yin\affmark[1], Di Huang\affmark[2], Hongyu Yang\affmark[2], Zehua Fu\affmark[1], Yunhong Wang\affmark[2], Liming Chen\affmark[1]\\
\affmark[1]Department of Mathematics and Informatics, Ecole Centrale de Lyon, Lyon, 69134, France\\
\affmark[2]School of Computer Science and Engineering, Beihang University, Beijing, 100191,China\\
\tt\small{\{yin.xiangnan,Liming.chen, zehua.fu\}@ec-lyon.fr}, \tt\small{\{dhuang,hongyuyang,yhwang\}@buaa.edu.cn}
}



\maketitle
\thispagestyle{empty}

\begin{abstract}
   The existing auto-encoder based face pose editing methods primarily focus on modeling the identity preserving ability during pose synthesis, but are less able to preserve the image style properly, which refers to the color, brightness, saturation, etc. In this paper, we take advantage of the well-known frontal/profile optical illusion and present a novel two-stage approach to solve the aforementioned dilemma, where the task of face pose manipulation is cast into face inpainting. By selectively sampling pixels from the input face and slightly adjust their relative locations with the proposed ``Pixel Attention Sampling" module, the face editing result faithfully keeps the identity information as well as the image style unchanged. By leveraging high-dimensional embedding at the inpainting stage, finer details are generated. Further, with the 3D facial landmarks as guidance, our method is able to manipulate face pose in three degrees of freedom, i.e., yaw, pitch, and roll, resulting in more flexible face pose editing than merely controlling the yaw angle as usually achieved by the current state-of-the-art. Both the qualitative and quantitative evaluations validate the superiority of the proposed approach.
\end{abstract}

\let\thefootnote\relax\footnotetext{\mycopyrightnotice}

\section{Introduction}

Face pose editing aims to change the pose of an input face image while keeping its original identity unchanged. It has many potential applications, \textit{e.g.},  face recognition, movie industry and entertainment. The current state-of-the-art has featured two main research lines in this field, \textit{i.e.},  3D reconstruction-based, and simple 2D based. 

For 3D reconstruction-based approaches, face pose editing is achieved by either mapping the 2D face images to 3D face models with fixed or regressed parameters~\cite{liu2018joint,zhao2017dual,yin2017towards} or directly regressing the UV map~\cite{deng2018uv,feng2018joint} of the input face. The advantage of such models is that pose control is not demanding. With the reconstructed 3D face, face images at any target pose can be obtained by 3D geometrical transformation and 2D projection. However, regressing either the parameters of predefined 3D models or the UV map requires large amounts of high-quality training data. Moreover, due to the restriction of the predefined model and the missing texture of extreme poses, fine details of the images are ignored. As a result, the faces generated by these approaches are generally not photo-realistic enough and require further refinements\cite{gecer2018semi}.

 Thanks to the development of Generative Adversarial Networks (GAN)~\cite{goodfellow2014generative}, a number of GAN based 2D approaches to face pose editing have been proposed in recent years. GAN has achieved great success in face image inpainting and facial attribute editing~\cite{cai2019fcsr,huang2018multimodal,pumarola2018ganimation,portenier2018faceshop,choi2018stargan}. However, the existing methods are generally only capable of editing the subtle attributes or local regions of the image, whereas the global structure remains almost unchanged. Regarding face pose manipulation, when changing the view angle from side to front, not only the local texture but also the global shape of the face image dramatically changes. Despite these difficulties, there still exist significant efforts tackling this problem  ~\cite{tran2017disentangled, huang2017beyond, hu2018pose, tian2018cr, qian2019unsupervised}. Most of the methods are implemented by an encoder-decoder structured network, with a bottleneck layer in the middle, where the faces are first encoded into a low dimensional feature vector, and then decoded into the image space conditioned by the pose information, \textit{e.g.}, CR-GAN~\cite{tian2018cr}, DR-GAN~\cite{tran2017disentangled}. However, there exists an intrinsic trade-off between the image style conserving capability and the identity preserving ability in the compact deep feature space, \textit{i.e.}, it is hard to model the expertise of both the face identity and other image properties, such as lightning condition, saturation, background color, \textit{etc.}

\begin{figure}[t]
\begin{flushleft}
   \includegraphics[width=\linewidth]{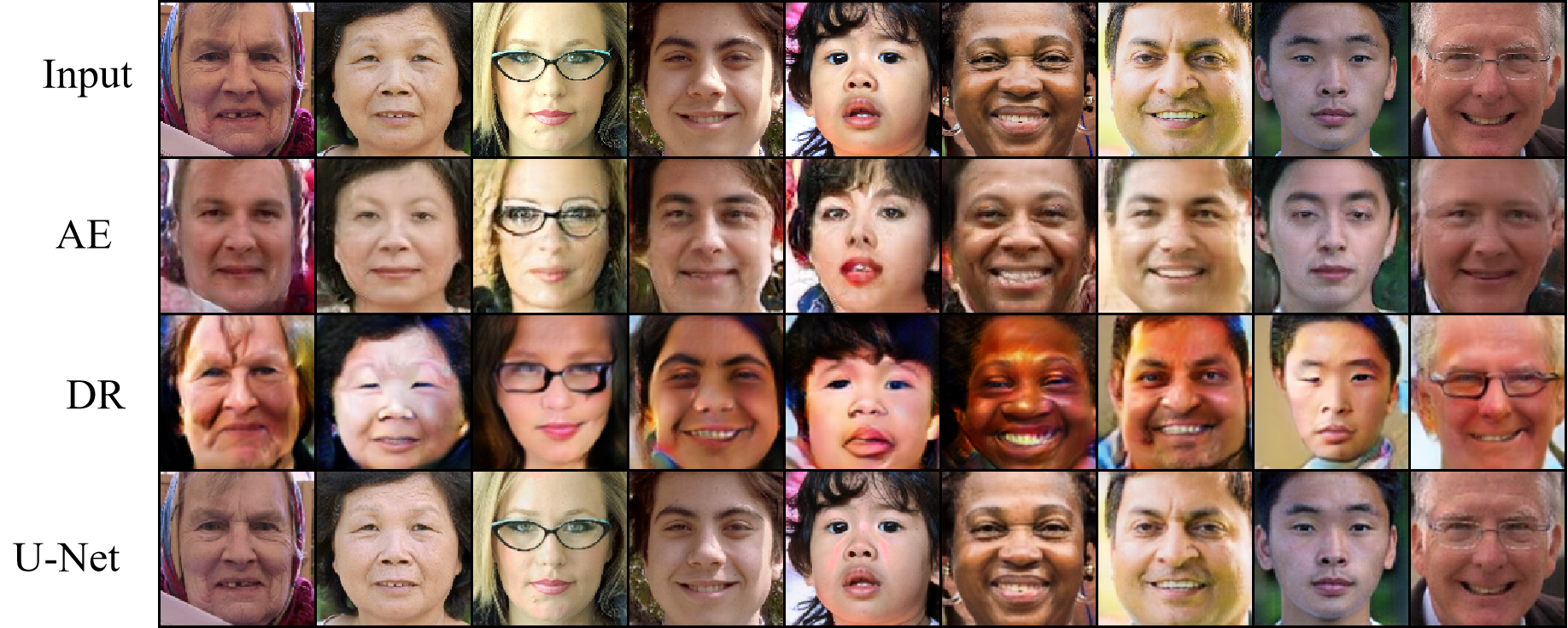}
   \end{flushleft}
   \caption{Illustration of the trade-off between identity preserving and style preserving.
}

\label{fig:structure comparison}
\end{figure}

To highlight the aforementioned dilemma that commonly incurs in current 2D based methods, we remove the face classification branch of DR-GAN~\cite{tran2017disentangled} (with the latent feature dimensionality of 320) and train the model only with the adversarial loss and the reconstruction loss. In this case, an adversarial auto-encoder (AE) is achieved, where the reconstruction loss aims to efficiently preserve the style of the input image, and the adversarial loss enforces the generated images photo-realistic. Figure~\ref{fig:structure comparison} illustrates the input images (the first row) and the results obtained by the adversarial auto-encoder (second row) and DR-GAN (third row), respectively. As can be seen, the auto-encoder properly preserves the style of the input image, but it fails maintaining the identities. The reconstructed faces by DR-GAN successfully catch the identity characteristics of the input images, whereas the output ones are distorted and present obvious artifacts. If it is even painful for the model to faithfully rebuild the given input face in terms of both style and the identity without any pose manipulation, how can we further expect it to preserve them after changing the pose? 

To fight the trade-off incurred by the low-dimensional restriction in the feature space, we seek solutions from the high-dimensional embeddings. But to make the condition label not ignored by the decoder, the encoding dimension should not be simply increased. The classical structure of U-Net~\cite{ronneberger2015u}, which adds skip-connections between symmetric layers of the encoder and the decoder, is able to prevent the problem of over-compression by concatenating features from the shallow layers in the reconstruction path. The last row of Figure ~\ref{fig:structure comparison} shows the corresponding reconstruction results, where both the identity and the style of the input face is well preserved. High-dimensional embedding is indeed promising in image synthesis, however, structures like U-Net convey too much low-level details, making it much more challenging to edit the face pose than on the low-dimension features, especially for the extreme shape changes. Therefore, how to enable face pose editing in the high dimensional feature space is the main problem to be solved. 

\begin{figure}[t]
    \centering
    
  \subfloat[\label{fig:opticalIllusion}]{%
  \begin{tabular}[b]{@{}c@{}}
       \includegraphics[width=0.2\linewidth]{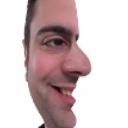}
      \hspace{1.5cm}
        \includegraphics[width=0.2\linewidth]{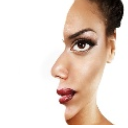}
    \end{tabular}}
    \\
    \subfloat[\label{fig:ambiguity}]{
        \includegraphics[width=0.7\linewidth]{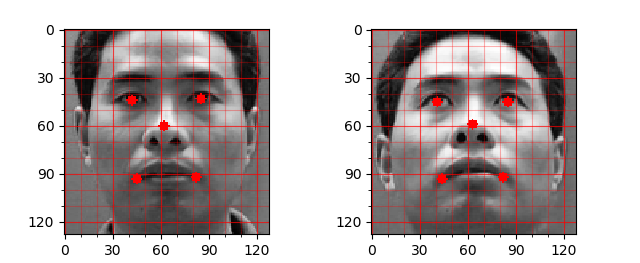}}
  \caption{(a) Example of front/profile optical illusion. Indicating that face images in different view angles still share pixel-level similarities. (b) The ambiguity of representing 3D face pose by 2D landmarks. The two faces above have almost the same landmark distribution, but are in different poses.}
  \label{fig:1} 
\end{figure}

To tackle the challenge above, we present a novel two-stage method and a module named ``{\bf P}ixel {\bf A}ttention {\bf S}ampling'' (PAS) in this paper. Inspired by the fact that face images of different view angles also share a large number of similar pixels as highlighted by the optical illusion of face images~\cite{opticalIllusion,opticalIllusion2} in Figure~\ref{fig:opticalIllusion}, we believe that these pixels are significant to construct the texture of a face image in the target view through sampling. Specifically, given a target pose, this PAS module selects pixels from the input image and slightly change their relative locations in a learning manner to match the target pose (similarly to a non-linear image warping). Thus the recovered face editing result possesses the target pose and shares the original texture simultaneously, faithfully keeping the identity information and image style unchanged. Due to the lack of texture in invisible regions, the results of PAS would possibly contain noises and holes, then the main task can be cast as image inpainting, which has been extensively studied. We feed the intermediate pose-edited face image into the aforementioned U-Net, so that the noises can be filtered out and holes filled. By incorporating the module of PAS, the low-level details preserved by the U-Net are no longer burdensome for the task of pose editing, instead, they become useful information for generating the visually compelling face images.

Further, by introducing the 3D landmarks rather than 2D ones to represent the head pose more precisely, we achieve a better flexibility of pose manipulation. On the contrary, the traditional methods like DR-GAN and CR-GAN merely manipulate face images in several discrete yaw angles, and TP-GAN~\cite{huang2017beyond} can only frontalize face images. Although CAPG-GAN~\cite{hu2018pose} uses 2D landmarks to guide the generation, it cannot generate faces in arbitrary poses as it claims, since using 2D landmarks to represent 3D angles can bring ambiguity as Figure~\ref{fig:ambiguity} shows. Besides, the 3D landmarks tends to provide richer shape-related information, further facilitating the synthesis of face images. 

In summary, our main contributions are as follows:
\begin{itemize}
    \item A novel two-stage face pose editing method is proposed, which casts the task of face pose manipulation as face inpainting, thereby enabling it fully utilize the fine details of the given input image by exploiting high-dimensional embedding.
    \item A new ``{\bf P}ixel {\bf A}ttention {\bf S}ampling'' module is designed, which effectively resolves the conflict between the identity and style preserving.
    \item The 3D facial landmarks is introduced to represent face poses for the first time, resulting in more flexible pose editing than using the discrete one-hot pose label or ambiguous 2D facial landmarks.
    \item The proposed method demonstrates competitive performance in comparison with the current state-of-the-art, both qualitatively and quantitatively. 
\end{itemize}
\section{Related Work}
\subsection{Generative Adversarial Network (GAN)}
In recent years, Generative Adversarial Networks (GAN) has been one of the most popular research directions for image generation. Traditional GAN is composed of a generator and a discriminator. The training follows an adversarial paradigm. To overcome the problems of unstable gradient and mode collapse, Wasserstein GAN (WGAN)~\cite{arjovsky2017wasserstein} proposes the earth move distance as metric in the discriminator's loss function. To enforce the Lipschitz constraint of the discriminator, SN-GAN~\cite{miyato2018spectral} applies spectral normalization to the weight parameters. Due to its simplicity and promising effect, most of the recent GAN based algorithms make use of this technique, including SN-GAN~\cite{zhang2018self}, BigGANs~\cite{brock2018large}, StyleGAN~\cite{karras2019style}, \textit{etc}. In our method, SN-GAN is also adopted in the structure.

\subsection{Image-to-Image Translation}
The combination of auto-encoder with discriminator has achieved impressive results in image-to-image translation~\cite{choi2018stargan,zhu2017unpaired,zhang2017age,pumarola2018ganimation}. In multi-domain image translation tasks, the domain information is provided either to the bottleneck layer of the auto-encoder~\cite{zhang2017age,he2019attgan,tran2017disentangled}, or to the entry of the encoder/generator~\cite{choi2018stargan,pumarola2018ganimation,hu2018pose}, by simply concatenating the domain label with the features or input images. Conditional batch normalization~\cite{de2017modulating} and conditional instance normalization (CIN)~\cite{dumoulin2016learned} provide another way of introducing the conditional label in addition to concatenation, via predicting the affine parameters of the normalized feature map (either by batch normalization or by instance normalization) from the input label. Here, the CIN technique is exploited in our decoder to avoid the operation of duplicating the label. 

In multi-domain image translation, the discriminator is used to not only estimate the image quality, but also control the target domain of the generated image. Our approach borrows the idea of projection discriminator~\cite{miyato2018cgans}, which introduces the reality score and the inner product of the embedded label with the features of the input data. 

\subsection{Face Pose Manipulation}
The existing methods can be roughly divided into two categories: 3D reconstruction based, and simple 2D based. 

For the 3D based models, DA-GAN~\cite{zhao2017dual} uses a predefined 3D face model to produce the synthesized faces with arbitrary poses, and the dual agents serve to keep the identity information stable and improve the realism, Feng \textit{et al.}~\cite{feng2018joint} train a model to regress the UV map from a single 2D image directly, which records the 3D shape information. Tran \textit{et al.}~\cite{tran2019learning} proposes a framework to learn a nonlinear 3DMM model from a large set of unconstrained face images. FF-GAN~\cite{yin2017towards} incorporates 3DMM~\cite{blanz1999morphable} into the GAN based structure, where the 3DMM coefficients provide the low-frequency information,  while the input image injects high-frequency local information.

For 2D based models, DR-GAN ~\cite{tran2017disentangled} learns a disentangled representation of face identity with the supervision of an auxiliary face classifier of the discriminator. TP-GAN ~\cite{huang2017beyond} employs a two-pathway architecture to preserve both global and local texture information separately, and generates the frontalized face images. With the guidance of 2D facial landmarks, CAPG-GAN~\cite{hu2018pose} is able to generate faces of arbitrary poses, where the couple-agent discriminator distinguishes the generated face/landmark pairs and profile/front pairs from ground-truth pairs, such design enables the algorithm generate face images of target poses while keeping the identity unchanged. CR-GAN ~\cite{tian2018cr} trains the generator to produce face images directly from the noises, together with the training of pose manipulation, maintaining the completeness of the learned embedding space. FNM~\cite{qian2019unsupervised} employs unsupervised training and synthesizes normalized face images of Multi-PIE~\cite{gross2010multi} style. Most of the above methods only focus on modeling the identity preserving ability, whereas they generally ignore the image style preserving ability, such as color, facial expression, lightning, \textit{etc.} Although it is claimed that the synthesized frontal face images improve the face verification accuracy, the generated face images are visually far from the input images, thus greatly limits their further usage scenarios other than face recognition.\cite{zhang2018face} frontalizes the face image by predicting the pixel displacement. However, it's hard to extend to the arbitrary face pose editing problem due to the time consuming SIFT feature extraction.

\section{Method}
\begin{figure}[t]
    \centering
  \subfloat[$I$\label{fig:variables:a}]{%
       \includegraphics[width=0.15\linewidth]{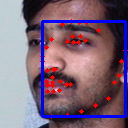}}
    \hfill
  \subfloat[$ldmk_I$\label{fig:variables:b}]{%
        \includegraphics[width=0.15\linewidth]{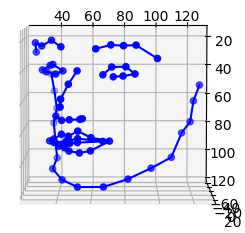}}
    \hfill
  \subfloat[$I_{tf}$\label{fig:variables:c}]{%
        \includegraphics[width=0.15\linewidth]{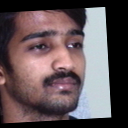}}
    \hfill
  \subfloat[$I_{seg\_tf}$\label{fig:variables:d}]{%
        \includegraphics[width=0.15\linewidth]{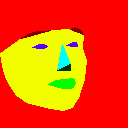}}
  \\
  \subfloat[$J$\label{fig:variables:e}]{%
        \includegraphics[width=0.15\linewidth]{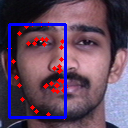}}
    \hfill
  \subfloat[$ldmk_J$\label{fig:variables:f}]{%
        \includegraphics[width=0.15\linewidth]{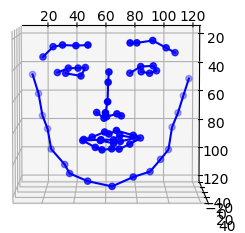}}
    \hfill
  \subfloat[blend\label{fig:variables:g}]{%
        \includegraphics[width=0.15\linewidth]{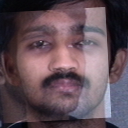}}
    \hfill
  \subfloat[$J_{seg}$\label{fig:variables:h}]{%
        \includegraphics[width=0.15\linewidth]{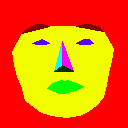}}
    \vspace*{3mm}
  \caption{(a) and (e) are the source image and the target one, respectively, where the corresponding landmarks and the bounding boxes of their bigger side are shown to illustrate our aligning strategy. (b) and (f) are the 3D facial landmarks detected by~\cite{bulat2017far}. (c) shows the aligned image $I_{tf}$. (g) is the alpha blend of $I_{tf}$ and $J$, illustrating that the target image shares pixel-level similarities with the source image. (d) and (h) are the segmentation maps of $I_{tf}$ and $J$ transformed from their 2D facial landmarks.}
  \label{fig:variables} 
\end{figure}

The goal of our method is to keep not only the identity but also the image style during face pose manipulation. We first define several notations: $(I, J)$ denotes paired face images in the training set, where $I$ is the source image, and $J$ is the target one. The 3D facial landmarks are denoted as $ldmk_I$ and $ldmk_J$, which could be detected by an off-the-shelf 2D$\backslash$3D facial landmark detector~\cite{bulat2017far}. $I_{tf}$ represents the input image after similarity transformation. To guide the training, the landmark based segmentation maps of $I_{tf}$ and $J$ are also required, which we denote as $I_{seg\_tf}$ and $J_{seg}$. These notations are visualized in Figure~\ref{fig:variables}.  

Our approach is composed of three major steps: preprocessing, pixel attention sampling, and image inpainting. They are described in detail subsequently.

\subsection{Preprocessing}

Given the fact that human faces are roughly left-right symmetrical, thus a face at an arbitrary pose always has at least one side fully exposed to the camera. This preprocessing step aims to align the fully exposed side of face $I$ to that of a target face $J$. 

The inputs of this step are the input face image $I$, its 3D facial landmarks $ldmk_I$, and the landmarks $ldmk_J$ of the image $J$ at a target pose. We first find the fully exposed side by calculating the bounding box region of the projected facial landmarks, as illustrated in Figure~\ref{fig:variables:a} and Figure~\ref{fig:variables:e}. Then, the least square regression on the corresponding landmarks is applied to calculate the transformation matrix, based on which the aligned image $I_{tf}$ could be obtained, as shown in Figure~\ref{fig:variables:c}.  From Figure~\ref{fig:variables:g}, we observe that $I_{tf}$ and $J$ indeed share pixel-level similarities. Finally, with the 2D facial landmarks of $I$ and  the transformation matrix obtained above, we obtain the 2D facial landmarks of $I_{tf}$, as well as the landmark based segmentation map $I_{seg\_tf}$. Besides, to guide the training process of the PAS module, the segmentation map of the target image $J_{seg}$ is also prepared at this stage.

\subsection{Pixel Attention Sampling}
\begin{figure*}
\begin{center}
\includegraphics[width=0.7\linewidth]{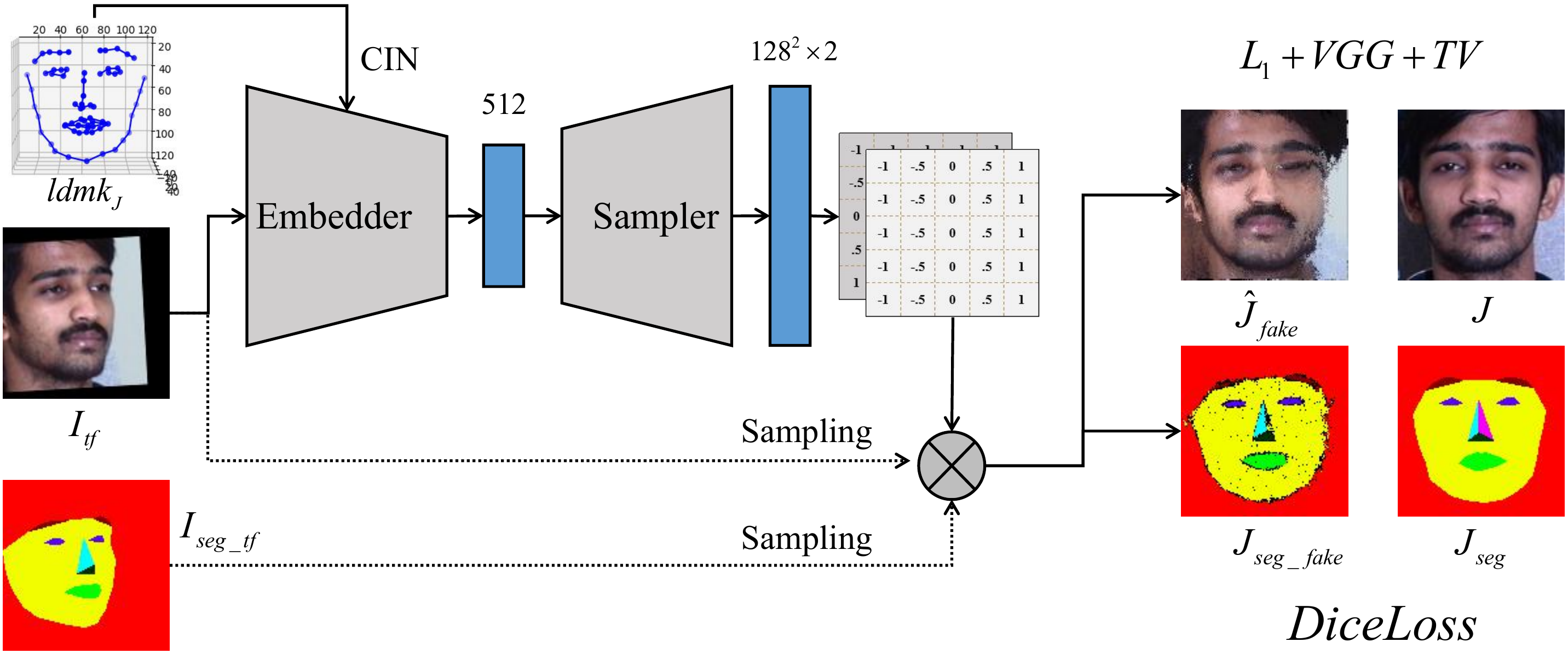}
\end{center}
   \caption{Structure of the proposed {\bf P}ixel {\bf A}ttention {\bf S}ampling (PAS) module.}
\label{fig:pas}
\end{figure*}

The previous preprocessing step delivers the input face image with the larger side aligned to the target pose. Despite the fact that the transformed input face image $I_{tf}$ and the face at the target pose $J$ share many similarities in terms of texture, there still exist great gaps between them, from the global shape to the finer details of textures. Therefore, our goal at this stage is to preserve and fine-tune their similar face regions while eliminating the major differences. This is achieved by a novel pixel sampling based module, which we call {\bf P}ixel {\bf A}ttention {\bf S}ampling module (PAS), since the process of sampling mainly ``focuses" on bridging the gaps. Figure~\ref{fig:pas} depicts the corresponding diagram.

Specifically, given the transformed image $I_{tf}$ and the target pose $ldmk_{J}$, PAS generates a two-channel coordinate sampling map of the same size as $I_{tf}$. The first channel holds the abscissa while the second one for the ordinate. Each pixel location of the map is registered a coordinate, indicating which input pixel of $I_{tf}$ that location will sample from. Note, the original pixel indices are converted into decimal coordinates ranging from -1 to 1, for the purpose of gradient backpropagation, and the final sampling is achieved by interpolating the adjacent pixels. Our sampling map is similar to the one used in the spatial transformer network\cite{jaderberg2015spatial}. The difference lies in that the one in \cite{jaderberg2015spatial} is determined by a 2D affine transform matrix, with only six parameters, whereas our sampling map is directly predicted by the neural network, resulting in $height\times width \times 2$ parameters in total. The PAS module is composed of two parts \textit{i.e.}, the image embedder and the sampler. The embedder consists of stacked convolution layers, conditional instance normalization~\cite{dumoulin2016learned} layers (CIN), and self-attention~\cite{zhang2018self} layers (SA). The CIN layers incorporates the 3D facial landmarks of the target pose to guide the embedding, and the SA layers enable the embedder focus more on the global structure of the face.The embedder finally outputs a 512-dimensional feature vector, which is further fed into the sampler to generate the sampling maps. The sampler is composed of fully connected layers and ReLU layers. After applying the obtained sampling map to the transformed input face image $I_{tf}$ and its corresponding segmentation map $I_{seg\_tf}$, we could obtain the intermediate face image at the target pose, denoted as $\hat{J}_{fake}$, and its corresponding fake segmentation map, denoted as $J_{seg\_fake}$. In order to maintain the reconstruction ability of the module, the original image $I$ and its corresponding 3D landmarks $ldmk_I$ are also fed into the PAS module and the reconstructed output $\hat{I}_{recon}$ is achieved.

The training process of the PAS is guided by the following losses:

\textbf{Pixel-wise loss} between $\hat{J}_{fake}$ and $J$, $\hat{I}_{recon}$ and $I$, which is commonly used in the image-to-image translation algorithms. It can be formulated as: 
 
 \begin{equation}
     \label{lpix}
     L_{pix} = L_1(\hat{J}_{fake}, J) + 0.1\cdot L_1(\hat{I}_{recon}, I)
 \end{equation}
where 
\begin{equation}
\label{l1}
    L_1(I, J) = \frac{1}{WHC}\sum_{x,y,c=1}^{W,H,C} 
    \left|I(x,y,c)-J(x,y,c)\right|
\end{equation}

Since it does not take much effort to learn an identity mapping, we set the weight of the reconstruction loss to 0.1, which makes the PAS module concentrate much more on the pose manipulation task.

\textbf{Segmentation loss} between $J_{seg\_fake}$ and $J_{seg}$. Based on the assumption that if the sampled image $\hat{J}_{fake}$ is close to the target image $J$, the segmentation map $J_{seg\_fake}$ should be close to the target segmentation map $J_{seg}$ as well. We therefore introduce a segmentation-related loss so as to push $\hat{J}_{fake}$ close to $J$. To facilitate the training converge, the segmentation loss is used as a complement to the aforementioned pixel-wise loss $L_{pix}$. Here, we make use of the Dice loss\cite{milletari2016v}, which has been widely exploited in image segmentation tasks, and it can be formulated as: 
\begin{equation}
\label{lseg}
    L_{seg} = \sum_{c=1}^N 1-\frac{2\sum\limits_{x,y} J_{seg}^c(x, y)\cdot J_{seg\_fake}^c(x,y)}{\sum\limits_{x,y} J_{seg}^c(x, y)+\sum\limits_{x,y} J_{seg\_fake}^c(x, y)}
\end{equation}
where $c$ represents the different classes of facial attributes. Since each pixel location $(x, y)$ of the segmentation map is represented by a $c$-dimensional one-hot vector, the fraction in Equation~\ref{lseg} is thus a simple intersection over union. The benefit of the adopted loss function is that it is independent to the amount of pixels of different classes.

\textbf{Perceptual loss}\cite{johnson2016perceptual} between $\hat{J}_{fake}$ and $J$. Perceptual loss is significant to preserve the identity information and high-level semantic features of the face images. We follow the work of \cite{zakharov2019few} and employ the pre-trained VGG-Face~\cite{Parkhi15} network to extract the features: 
\begin{equation}
\label{lper}
    L_{per} = VGG_{loss}(\hat{J}_{fake}, J)
\end{equation}
with
\begin{equation}
\begin{aligned}
\label{vgg}
    VGG_{loss}(I, J) =
    \sum_{i}\left|VGG_{Face}(I)_i-VGG_{Face}(J)_i\right|
\end{aligned}
\end{equation}
 where $i$ is the layer index of the pre-trained model and $i\in$ \{3,8,15,22,29\}, which are the last convolutional layer of each feature map scale.

\textbf{Total variation loss}. Total variation~\cite{mahendran2015understanding} loss has been widely used in GAN based algorithms for its powerful ability of reducing the noises and smoothing the generated results. In our PAS module, there inevitably exist obvious noises, since the resultant face image is pixel-wise sampled from the transformed input face image. Therefore, the TV loss is incorporated: 
\begin{equation}
\label{ltv}
    L_{tv} = TV(\hat{J}_{fake}) + TV(\hat{I}_{recon})
\end{equation}
where
\begin{equation}
\begin{aligned}
\label{tv}
    TV(I) &=
     \sum_{x,y,c=1}^{W-1,H,C}\left|I(x+1,y,c)-I(x,y,c)\right|^2 \\
      &+\sum_{x,y,c=1}^{W,H-1,C}\left|I(x,y+1,c)-I(x,y,c)\right|^2
\end{aligned}
\end{equation}

The overall training loss of the PAS module is a sum of the above losses:
\begin{equation}
    \label{lall}
    L_{sampler} = L_{pix} + L_{seg} +L_{per} + L_{tv}
\end{equation}

\begin{figure*}[t]
\begin{center}
   \includegraphics[width=0.8\linewidth]{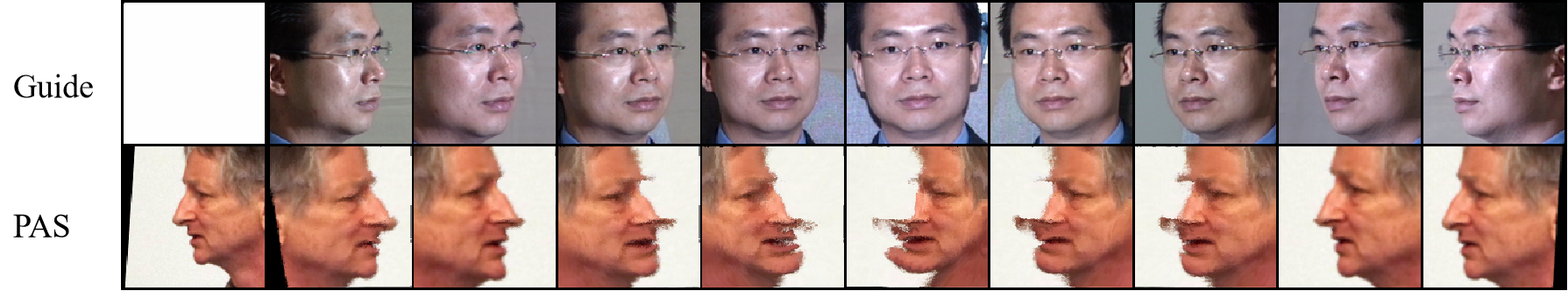}
\end{center}
   \caption{The result of PAS. The first row shows the guiding face images at target poses, the first image in the second row is the input face image, and the remaining images are synthesized faces based on the landmarks of the guiding face images. We can see that the pixels are sampled and adjusted to the target pose. The noises and holes will be removed or filled at the next image inpainting stage. 
}
\label{fig:sampling_result}
\end{figure*}

Thanks to the PAS module, we achieve a face image whose facial attributes have been aligned to the target pose location, with the original identity and style characteristics well preserved. It should be noted that, as the sampling is accomplished by interpolating adjacent pixels, it only modifies the location of the pixels within a small area around them, the PAS module is thus not able to sample for instance the left eye from the right one or the opposite. As a result, the sampled face images possibly contain artifacts, holes and noises, as illustrated in Figure~\ref{fig:sampling_result}.  In order to further improve the generated image quality, image inpainting is introduced subsequently.

\subsection{Image Inpainting}
The image inpainting stage is to restore the holes and remove the noises and artifacts on the intermediate faces generated by PAS, and finally generate photo-realistic face images. To accomplish this goal, we introduce a Conditional Adversarial Auto-Encoder, where the discriminator is implemented by a projection discriminator~\cite{miyato2018cgans}, and the auto-encoder is based on the U-Net structure~\cite{ronneberger2015u}. We also make use of CIN layer to merge the information provided by 3D facial landmarks, the identity features, and the image features, thereby making the generated face image in desired pose and shape. More precisely, the inputs of the encoder are the images generated by PAS together with their target poses, \textit{i.e.}, $\hat{J}_{fake}$ with $ldmk_J$ for the task of pose manipulation, and $\hat{I}_{recon}$ with $ldmk_I$ for the task of reconstruction. To well preserve the face identity, the decoder is conditioned by the high level feature extracted by the pre-trained LightCNN~\cite{wu2018light} model, where the parameters of the fully connected layer is fine-tuned during training. The outputs of the auto-encoder are denoted as $J_{fake}$ and $I_{recon}$, whose ground truths are $J$ and $I$, respectively. To further improve the model's generalization ability, unpaired face images could also be exploited to supplement the training set. This is achieved by feeding the network with the partially occluded face images $S_{occ}$ and their 3D landmarks, and expect the network to output $S_{recon}$ restoring the original $S$. For the discriminator, we feed all the generated images, including $J_{fake}$, $I_{recon}$ and $S_{recon}$, as fake samples, while their corresponding ground truth as the genuine ones, with $dis_{real}$ and $dis_{fake}$ as output, respectively.

The loss function of the inpainting network is composed of four parts:

\textbf{Pixel-wise loss}, formulated as:
\begin{equation}
\label{lpix2}
\begin{aligned}
    L_{pix} = L_1(J_{fake}, J) + \lambda\cdot L_1(I_{recon}, I) + L_1(S_{recon},S)
\end{aligned}
\end{equation}
where $L_1$ is defined in equation~\ref{l1}.  

\textbf{Perceptual loss} to capture the semantic similarity:
\begin{equation}
\label{lper2}
    L_{per} = VGG_{loss}(J_{fake}, J) + VGG_{loss}(S_{recon}, S)
\end{equation}
where $VGG_{loss}$ is defined in equation~\ref{vgg}. We do not include $(I_{recon}, I)$ here, because compared to image reconstruction task, image inpainting and pose manipulation are more likely to lose the identity consistency.

\textbf{Identity loss} to maintain the identity-related characteristics stable. We use the pre-trained LightCNN~\cite{wu2018light} to extract the identity feature of the synthesized image and the target image, and minimize the $L_1$ loss of them:
\begin{equation}
\label{lid}
    L_{id} = \frac{1}{N}\sum_{i=1}^N\left| F(J_{fake})_i-F(J)_i\right |
\end{equation}

\textbf{Adversarial loss} to guarantee the generated image quality:
\begin{equation}
\label{gen_adv}
    L_{adv}=-dis_{fake}
\end{equation}

Besides, we also incorporate the total variation loss to reduce the spike artifacts. The overall training loss of the generator of the image inpainting network is a sum of the aforementioned losses:
\begin{equation}
\label{lgen}
    L_{gen} = L_{pix}+L_{per}+L_{id}+L_{tv}+L_{adv}
\end{equation}

Following the work of~\cite{lim2017geometric}, the {\bf discriminator loss} is defined as:
\begin{equation}
    L_{dis} = \mathbf{max}(1-dis_{real},0)+\mathbf{max}(1+dis_{fake}, 0)
\end{equation}

\section{Experiments}
Given an input face image, the proposed method aims to manipulate its pose while keeping the identity unchanged along with its style. Correspondingly, we evaluate it in two aspects: the style-conserving skill and the identity-preserving ability during face pose editing. In this section, we present the training details first, then the qualitative analysis for face style conserving, followed by the quantitative results for identity preserving. Ablation studies are also carried out to highlight the effectiveness of the proposed PAS module. 

\subsection{Training details}
The training is based on four databases: Multi-PIE~\cite{gross2010multi}, 300W-LP~\cite{zhu2016face}, CAS-PEAL-R1~\cite{gao2007cas}, and CelebA~\cite{liu2015deep}. \textbf{Multi-PIE} has four sessions with face images under 13 poses and 20 illuminations. We follow Setting 1 of TP-GAN~\cite{huang2017beyond} and train the proposed algorithm on the first 150 subjects of session 1, then test on the remained 99 subjects. \textbf{300W-LP} contains large-pose face images synthesized from 300W~\cite{sagonas2013300}. After manually filtering out the low-quality images, we have 40,159 images from 2,815 subjects in total.  \textbf{CAS-PEAL-R1} contains 1,040 subjects. For each subject, gray-scale images across 21 different poses are included. \textbf{CelebA} is a large-scale face attributes dataset with more than 200K celebrity images in it. 

During the training process, we use the occluded face images as input, and train the U-Net based generator to restore the original face images. This operation improves the generalization ability of the network, and make the generated images photo-realistic. All of the training images are cropped to 128$\times$128 pixels. The learning rate is set to $1\mathbf{e}^{-4}$, and the Adam~\cite{kingma2014adam} optimizer is utilized with betas of [0.9, 0.999]. We first pre-train the generator and the discriminator on CelebA for 20000 iterations, making it a fundamental image inpainting model, which facilitates the subsequent training procedure. Then, we train the proposed PAS model and the image inpainting model jointly for 110000 iterations in total. Observing that the CAS-PEAL-R1 dataset consists of gray-scale images, which degenerates the color saturation of the generated images, we thus exclud the data of CAS-PEAL-R1 for the last 10000 iterations.

\begin{figure*}[ht]
\begin{center}
   \includegraphics[height=6.3cm]{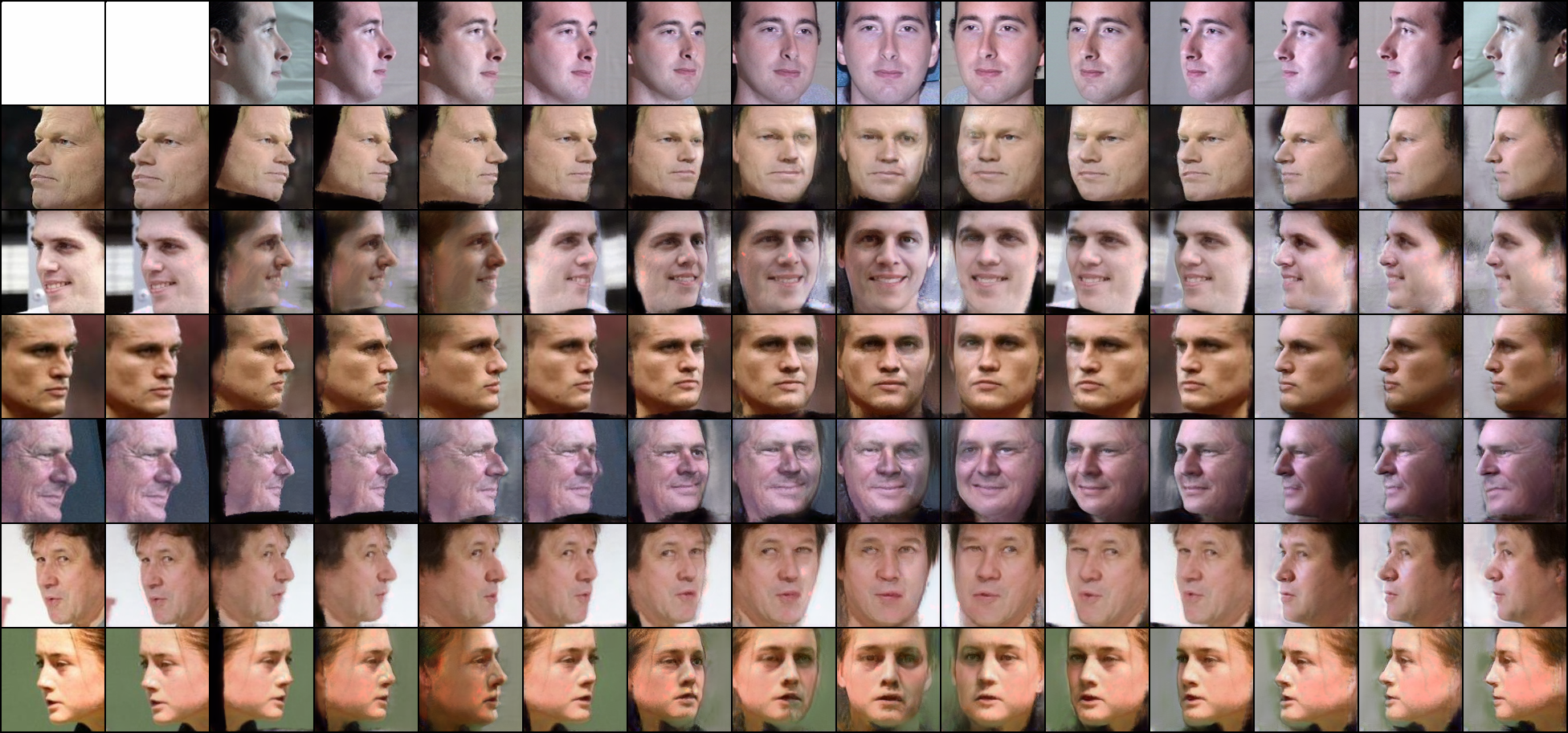}
\end{center}
   \caption{The final result of our approach. The first row shows the guidance images. The input images are in the first column, and the simple reconstructed images are in the second column. The rest images are the pose editing results based on the landmarks of the exemplars.
}

\label{fig:final}
\end{figure*}

\subsection{Style-conserving validation}
\begin{figure*}[t]
    \centering
  \subfloat[\label{fig:results2:a}]{%
       \includegraphics[height=4.5cm]{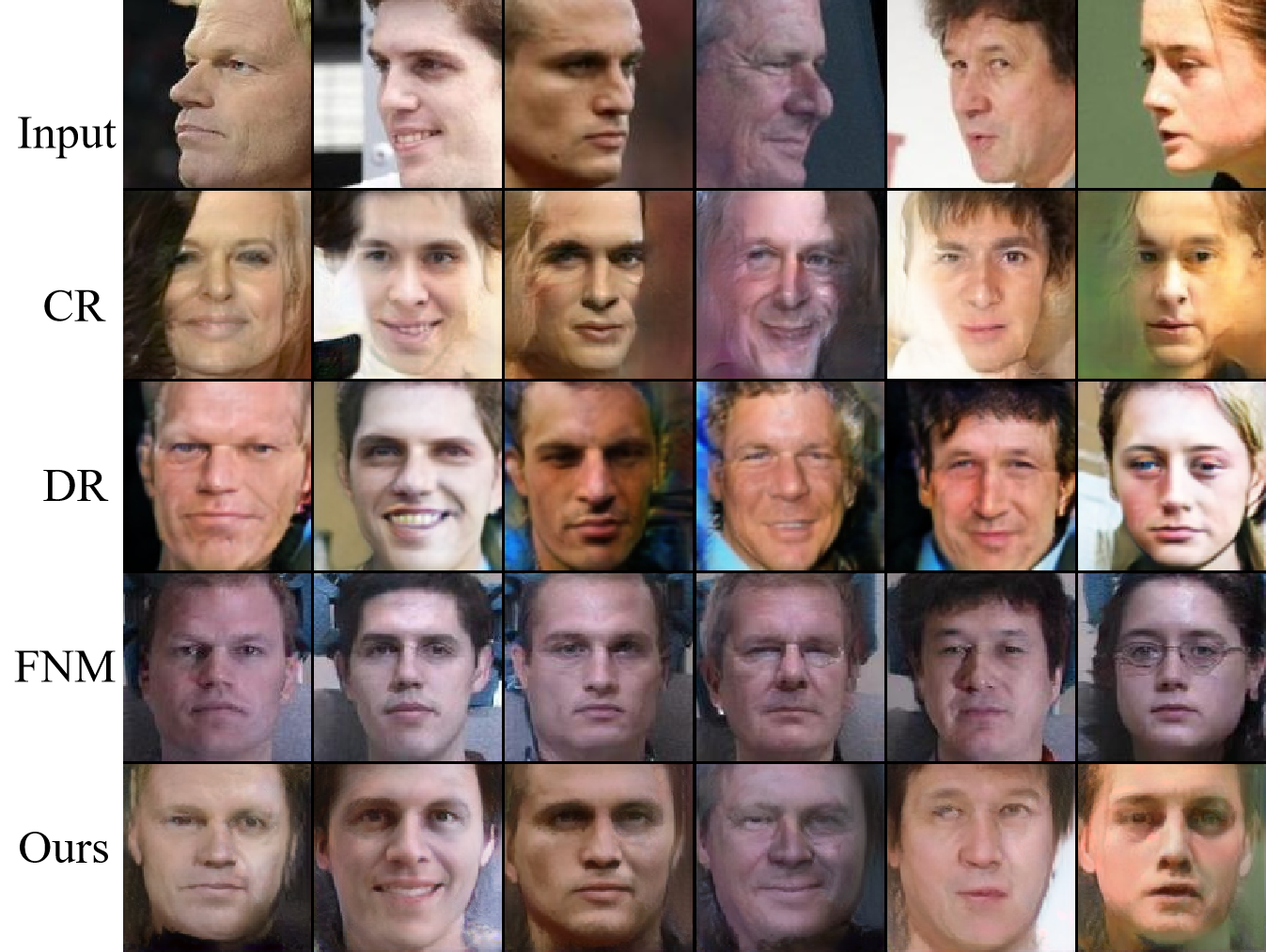}}
    \hfill
  \subfloat[\label{fig:results2:b}]{%
        \includegraphics[height=4.5cm]{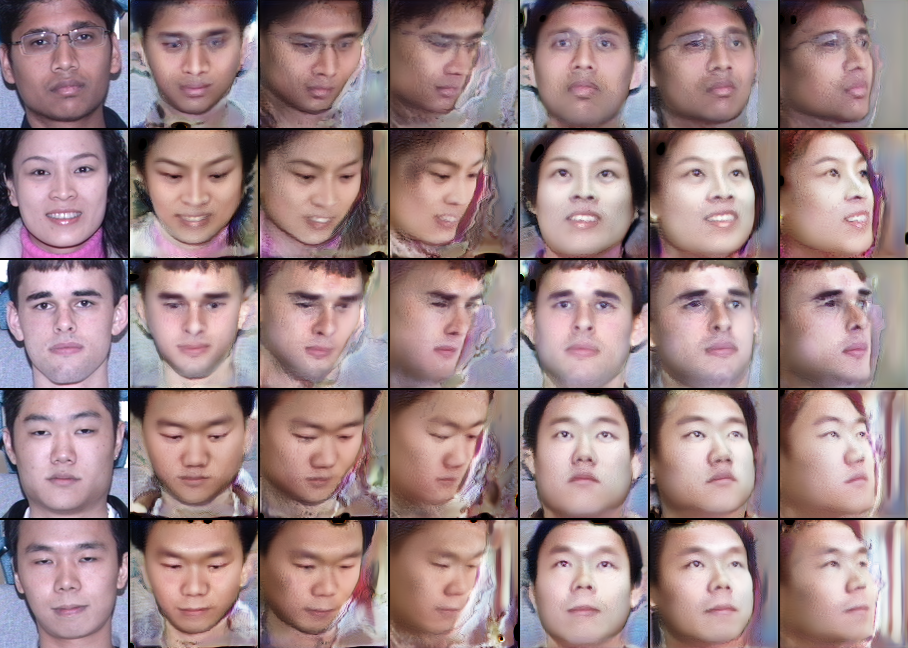}}
    \hfill
  \subfloat[\label{fig:results2:c}]{%
        \includegraphics[height=4.5cm]{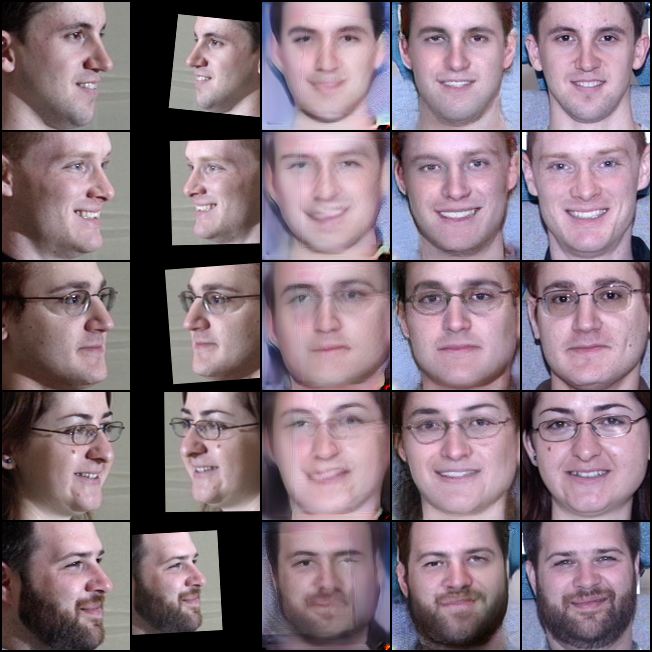}}
  \caption{(a) From top to bottom shows the input images, results of CR-GAN, DR-GAN, FNM and our method. (b) From left to right are the input images and the generated images with both yaw angles and pitch angles changed. (c) From left to right are the input images, half-face aligned images, frontalized images w/o PAS module, results of the proposed algorithm, and the ground-truth images.}
  \label{fig:results2} 
\end{figure*}

Multi-PIE images under different poses are used as the guiding images, the pose of faces from CelebA are edited accordingly. As shown in Figure~\ref{fig:final}, the synthesized face images comply with the guiding faces in term of pose. They are visually photo-realistic and both the identities and the styles are well preserved, clearly validating the effectiveness of the proposed method. A further qualitative comparison of our method and CR-GAN~\cite{tian2018cr}, DR-GAN~\cite{tran2017disentangled} and FNM~\cite{qian2019unsupervised} are demonstrated in Figure~\ref{fig:results2:a}. As can be seen, our results are more visually convincing and the styles are closer to the input images compared to DR-GAN, and the identities are better preserved than CR-GAN. As for FNM, the generated image style is more similar to the training set, where the lighting and expressions are normalized, and the color has been changed, by contrast, our results better preserve those characteristics of the input face images. Moreover, the proposed approach is able to manipulate face poses in three degrees of freedom, resulting in more flexible pose editing results than merely controlling the yaw angles as usually achieved by previous methods. Figure~\ref{fig:results2:b} shows the results of editing both pitch and yaw angles of input face images (leftmost).

Quantitative evaluations are further performed. We calculate the FID score of the above models, on the frontalized large pose face images from CelebA, the results are shown in Table~\ref{tab:fid}, indicating that the proposed method generates face images with styles closer to the input face images, which can be applied to more perceptual applications. 

\begin{table}[t]
\caption{FID score of frontalized face images (lower is better)}
    \centering
    \small
    \begin{tabular}{c c c c}
    \hline
     CR-GAN & DR-GAN & FNM & ours\\
    \hline
    \hline
    204 & 122 & 150 & 105\\
    \hline
    \end{tabular}
    \label{tab:fid}
\end{table}

\begin{table}[]
\caption{Rank-1 recognition rates (\%) across views, illuminations and emotions under Setting 1. }
    \centering
    \scriptsize
    \begin{tabular}{c c c c c c c}
    \hline
    Method & $\pm 90^\circ$ & $\pm 75^\circ$ & $\pm 60^\circ$ & $\pm 45^\circ$ & $\pm 30^\circ$ & $\pm 15^\circ$\\
    \hline
    \hline
    HPN~\cite{ding2017pose} & 29.82 & 47.57 & 61.24 & 72.77 & 78.26 & 84.23   \\
    c-CNN~\cite{xiong2015conditional} & 47.26 & 60.7 & 74.4 & 89 & 94.1 & 97.0 \\
    TP-GAN~\cite{huang2017beyond} & 64.0 & 84.1 & 92.9 & 98.6 & 99.9 & 99.8 \\
    PIM~\cite{zhao2018towards} & 75.0 & 91.2 & 97.7 & 98.3 & 99.4 & 99.9 \\
    CAPG-GAN~\cite{hu2018pose} & 77.1 & 87.4 & 93.7 & 98.3 & 99.4 & 99.9 \\
    FNM~\cite{qian2019unsupervised} & 55.8 & 81.3 & 93.7 & 98.2 & 99.5 & 99.9 \\
    \hline
    Light CNN~\cite{wu2018light} & 2.6 & 10.5 & 32.7 & 71.2 & 95.1 & 99.8 \\
    Ours& 45.5 & 78.7 & 90.0 & 99.6 & 99.9 & 100\\
    \hline
    \end{tabular}
    \label{tab:my_label}
\end{table}

\subsection{Identity-preserving ability evaluation}

There are 249 subjects in Session 1 of Multi-PIE. Following the Setting 1 of TP-GAN, we use the first 150 subjects for training, and the remaining 99 subjects for testing. The identity preserving ability is evaluated by Rank-1 recognition rate. The face with frontal view and normal illumination in the testing set compose the gallery, and the rest non-frontal images are used as probe.

The evaluation is conducted based on the features extracted by the pre-trained Light-CNN model. We directly extract the features of the probe images as baseline. For the proposed method, we first frontalize the probe face images, based on which their face representations are extracted. As can be seen from Table~\ref{tab:my_label}, the proposed method achieves similar or even better Rank-1 recognition rate in comparison with the baseline and state-of-the-art algorithms when the rotation angle is smaller than $60^\circ$. For larger rotation angles ($\geq 60^\circ$), the proposed algorithm drastically outperforms the baseline, whereas it does not perform as well as the SOTA algorithms. There exist two possible reasons: 1) The face images of extreme poses share relatively less pixels with the face images of front view, thus the pixels sampled by the PAS module are not sufficient enough for the following inpainting stage, and 2) most of the SOTA algorithms normalize the face images into a consistent style, where the information irrelevant to identity is filtered out, in contrast, our method preserves relatively more style information.

\subsection{Ablation Study}
To highlight the effectiveness of the PAS module, the ablation study is conducted by removing it and training the U-Net based conditional adversarial auto-encoder directly. For the sake of fair comparison, we apply the same preprocessing pipeline (\textit{i.e.}, align the larger side of the input image to match the target pose) and train the model with the same number of iterations. Figure~\ref{fig:results2:c} shows the results. As can be seen, the synthesized images without PAS are blurred. More specifically, in the second row and the fourth row, the mouths are not well aligned, and the unexpected edges of the aligned input images are not well removed. The results indicate that it is indeed difficult for the single U-Net based model to change the original patterns of the input image thus results in undesired artifacts.

\section{Conclusion}
In this work, we first carefully analyze the trade-off between the style-preserving ability and the identity-preserving ability of the existing 2D based pose manipulation methods. Based on the observation that face images in different poses share a large number of pixels, we propose a novel pose editing method and a sophisticatedly designed PAS module. The method selectively samples pixels from the input face and adjust their relative locations with the PAS module, so that the recovered face editing result match the target pose and faithfully keeps the original identity and style information unchanged. In this way, we convert the pose manipulation problem to a image inpainting problem, and further make the best of the finer details in the original face images to obtain convincing pose editing results. We also utilize 3D facial landmarks to represent the face pose, which is more precise and flexible comparing to the one-hot labels and the 2D facial landmarks adopted in previous studies. Extensive experiments validate that the proposed pose editing approach preserves the style information of the input images better than the existing methods. 

{\small
\bibliographystyle{ieee}
\bibliography{submission_example}
}

\end{document}